\documentclass[lettersize,journal]{IEEEtran}
\usepackage{amsmath,amsfonts}
\usepackage{algorithmic}
\usepackage{algorithm}
\usepackage{array}
\usepackage[caption=false,font=normalsize,labelfont=sf,textfont=sf]{subfig}
\usepackage{textcomp}
\usepackage{stfloats}
\usepackage{url}
\usepackage{verbatim}
\usepackage{graphicx}
\usepackage{cite}
\usepackage{multirow}
\usepackage{hhline}
\begin{document}

\title{CrossTrafficLLM: A Human-Centric Framework for Interpretable Traffic Intelligence via Large Language Model}

\author{Zeming Du, Qitan Shao, Hongfei Liu, Yong Zhang
        % <-this % stops a space
\thanks{Yong Zhang is the corresponding author.

Zeming Du, Yong Zhang, Qitan shao, Hongfei Liu are with Beijing Key Laboratory of Multimedia and Intelligent Software Technology, Beijing Artificial Intelligence Institute, Faculty of Information Technology, Beijing University of Technology, Beijing, China, 100124. (dzm18800157113@126.com; zhangyong2010@bjut.edu.cn; shaoqt@emails.bjut.edu.cn; fei837524919@emails.bjut.edu.cn}% <-this % stops a space
\thanks{This work was supported by National Natural Science Foundation of China(No.62472014, 72303221, U21B2038).
}}

\markboth{Journal of \LaTeX\ Class Files,~Vol.~14, No.~8, August~2021}%
{Shell \MakeLowercase{\textit{et al.}}: A Sample Article Using IEEEtran.cls for IEEE Journals}


\maketitle

\begin{abstract}
While accurate traffic forecasting is vital for Intelligent Transportation Systems (ITS), effectively communicating predicted conditions via natural language for human-centric decision support remains a challenge and is often handled separately. To address this, we propose CrossTrafficLLM, a novel GenAI-driven framework that simultaneously predicts future spatiotemporal traffic states and generates corresponding natural language descriptions, specifically targeting conditional abnormal event summaries. We tackle the core challenge of aligning quantitative traffic data with qualitative textual semantics by leveraging Large Language Models (LLMs) within a unified architecture. This design allows generative textual context to improve prediction accuracy while ensuring generated reports are directly informed by the forecast. Technically, a text-guided adaptive graph convolutional network is employed to effectively merge high-level semantic information with the traffic network structure. Evaluated on the BJTT dataset, CrossTrafficLLM demonstrably surpasses state-of-the-art methods in both traffic forecasting performance and text generation quality. By unifying prediction and description generation, CrossTrafficLLM delivers a more interpretable, and actionable approach to generative traffic intelligence, offering significant advantages for modern ITS applications.
\end{abstract}


\section{Introduction}
\IEEEPARstart{T}{raffic} forecasting is a fundamental component of intelligent transportation systems (ITS), underpinning a wide range of applications such as urban traffic planning, route navigation, congestion mitigation, and emergency response. Accurate prediction of traffic conditions is vital for city managers, travelers, and transportation service providers, enabling more efficient mobility and proactive management of road networks.

In recent years, deep learning techniques have pushed the boundaries of traffic forecasting. Models based on graph neural networks~\cite{stgcn}, spatiotemporal attention mechanisms~\cite{gman}, and transformer architectures~\cite{astgcn} have achieved impressive results in predicting traffic flows, speeds, and congestion patterns. These advances have led to improved accuracy and robustness, especially in capturing complex spatial and temporal dependencies within large-scale road networks.

Despite these successes, current traffic forecasting models predominantly focus on processing and predicting numerical data, such as traffic speed or flow. However, a critical gap remains in the transition from "computational intelligence" to "human-centric social intelligence." Current forecasting models predominantly focus on numerical accuracy, treating the system as a "black box" that outputs distinct spatiotemporal values (e.g., flow, speed). This numerical-centric paradigm ignores the \textit{why} and \textit{how} behind traffic dynamics—information often embedded in unstructured semantic signals ~\cite{chat} (e.g., social media reports, emergency logs) describing abnormal events. Consequently, these models struggle to provide actionable, interpretable insights for human decision-makers, creating a semantic barrier in human-AI collaboration.

\begin{figure}[!t]
\centering
\includegraphics[width=3.5in]{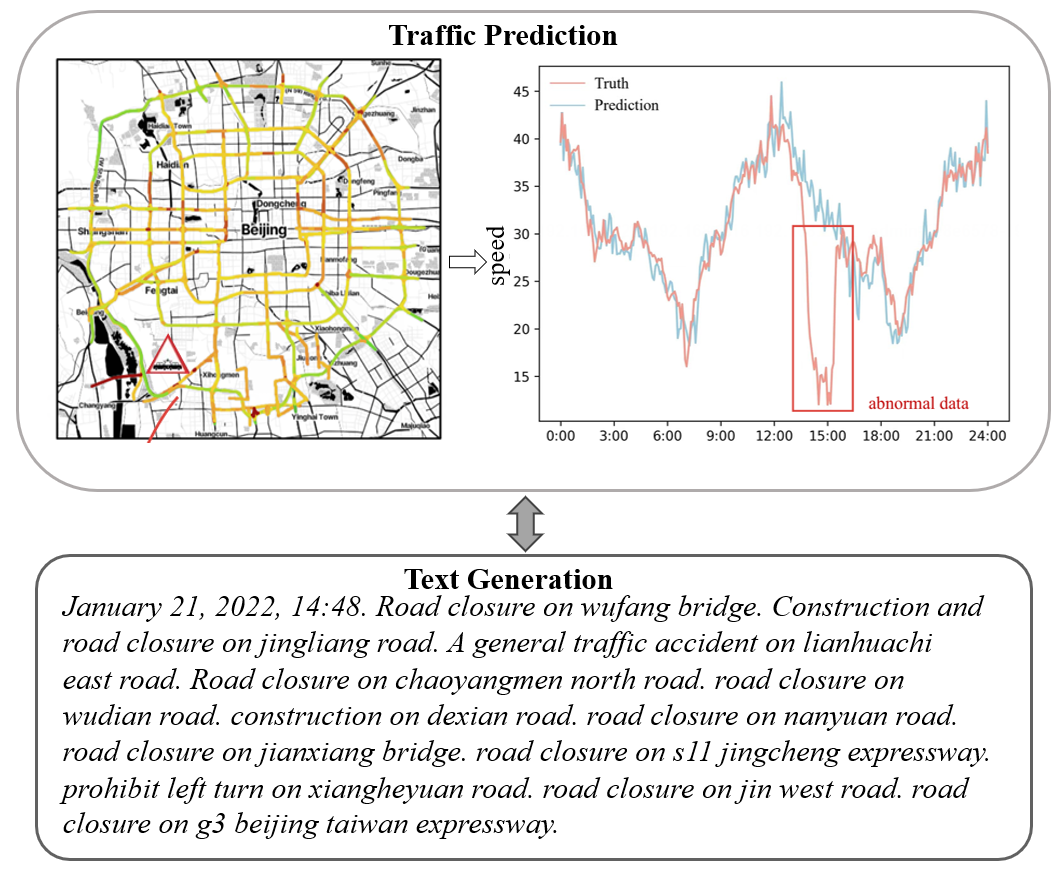}
\caption{An illustration of the benefits provided by this approach includes: 1) The model can learn from abnormal event texts to better recognize and anticipate non-routine traffic situations, thereby improving the accuracy of traffic forecasting under abnormal conditions. 2) The system is able to generate corresponding abnormal event descriptions, offering clear and intuitive explanations for complex or unexpected traffic states.}
\label{fig_1}
\end{figure}

Beyond their potential value for forecasting, textual descriptions of traffic conditions play a crucial role throughout modern transportation systems.  (as shown in Fig.~1) On the one hand, abnormal traffic events are often initially reported through unstructured text channels like traffic reports, social media posts, or navigation alerts before they appear in numerical data streams, providing early warnings and contextual information that could improve prediction accuracy. On the other hand, transportation managers and everyday users require interpretable, qualitative descriptions of traffic conditions to make informed decisions. While machines can efficiently process numerical predictions, humans benefit from natural language summaries that describe not just what is happening but also why it is occurring, especially during abnormal situations.

Nevertheless, integrating such textual data into traffic forecasting remains a significant challenge. Most existing forecasting frameworks primarily operate on structured numerical data and lack mechanisms to effectively incorporate unstructured text. This creates a disconnect between the rich contextual information available in textual reports and the quantitative predictions generated by current models. Additionally, most systems provide only numerical forecasts without translating these into human-readable insights, limiting their practical utility for non-technical users who need to understand the broader traffic situation.

Motivated by these observations, we aim to address the following key challenges: (1) How can unstructured textual information be effectively integrated to enhance the accuracy and interpretability of traffic forecasting, especially in non-recurrent or abnormal situations? (2) How can we construct a unified framework that not only predicts future traffic states but also generates meaningful textual summaries, thereby supporting both automated decision-making and human understanding in intelligent transportation systems?

The emergence of Generative AI (GenAI) and Large Language Models(LLMs)~\cite{gpt3, gpt4, deepseek, llama} offers a new opportunity to bridge this gap. These models demonstrate unprecedented capabilities in understanding and generating natural language, potentially enabling transportation systems to integrate textual reports about traffic events and generate contextual explanations alongside numerical predictions. To the best of our knowledge, no prior work has simultaneously realized both cross-modal traffic forecasting and automated traffic description generation within a unified LLM-enhanced framework.

In this paper, we introduce \textbf{CrossTrafficLLM}, a human-centric framework for interpretable traffic intelligence. Unlike previous works that treat text as static auxiliary features, we propose a generative alignment mechanism. Our approach uses LLM-driven semantic embeddings to dynamically reconfigure the traffic graph structure, allowing the "meaning" of an  event to physically alter how traffic is predicted.Our key contributions are:

1) \textbf{Unified Generative Framework}: We move beyond simple multi-task learning to propose a deeply coupled architecture where semantic context refines physical prediction, and physical states ground semantic generation.

2) \textbf{Semantic-Physical Alignment}: We introduce a text-guided adaptive graph convolutional network that maps unstructured event descriptions to specific road network topologies, resolving the challenge of cross-modal fusion.

3) \textbf{Interpretable Decision Support}: We demonstrate that CrossTrafficLLM surpasses SOTA methods in both tasks, providing a tool for GenAI-augmented social transportation systems.

The remainder of this paper is organized as follows: Section II reviews related work in traffic prediction, text generation, and multi-modal transportation models. Section III details the architecture and components of our CrossTrafficLLM framework. Section IV describes our experimental setup and evaluation metrics. Section V presents our results and comparative analysis. Finally, Section VI discusses implications, limitations, and directions for future research.

\section{Related Work}
In this section, we review the development of traffic forecasting methods, including the integration of external information sources, and then discuss recent advances in the application of large language models (LLMs) for traffic-related tasks.

\subsection{Traffic Forecasting: From Statistical Models to Multi-modal Integration}
Traffic forecasting methods have evolved significantly, beginning with classical statistical models such as ARIMA~\cite{arima} and Kalman filtering~\cite{kalman}, which captured temporal dependencies but struggled to represent the complex spatial-temporal dynamics of real-world traffic networks. The introduction of deep learning, particularly Recurrent Neural Networks (RNNs) and their variants such as LSTM~\cite{lstm} and GRU~\cite{gru}, improved temporal modeling but still faced limitations in capturing spatial correlations among road segments. Convolutional Neural Networks (CNNs)~\cite{cnn} addressed spatial feature extraction on grid-structured data, although they were less effective for non-Euclidean road networks.

Graph-based models, especially Graph Neural Networks (GNNs), further advanced the field by explicitly modeling road networks as graphs. Variants such as GCN~\cite{gcn}, DCRNN~\cite{dcrnn}, and STS-GCN~\cite{stsgcn} demonstrated the effectiveness of jointly learning spatial and temporal dependencies. More recently, attention-based and transformer architectures~\cite{gman, informer , crossformer , itransformer  , timexer} have set new performance benchmarks by capturing long-range spatial-temporal correlations.

Beyond purely numerical data, researchers have explored the integration of external information—including weather~\cite{weather}, social media~\cite{social-media}, and event data~\cite{event}—to enhance traffic prediction. Notably, studies have shown the effectiveness of social media content in assisting traffic forecasting and identifying abnormal events~\cite{twitter, context-aware, report1, report2, report3}. Multi-modal approaches such as MGNN~\cite{mgnn} fuse heterogeneous data sources, yet typically treat textual or contextual signals as auxiliary features or simple embeddings. This limits the full exploitation of rich semantic information present in unstructured text. Nevertheless, these efforts underscore the importance and potential of leveraging external variables—especially textual data—for improving the detection and interpretation of abnormal traffic states.

Despite these advances in multi-modal traffic forecasting, current approaches inadequately exploit unstructured textual information, typically treating text as simple auxiliary features rather than leveraging its rich semantic content. This limitation presents an opportunity to develop more sophisticated frameworks that effectively integrate natural language processing with traffic prediction, potentially enhancing both accuracy and interpretability while bridging the gap between structured traffic data and unstructured textual descriptions.

\subsection{Large Language Models for Traffic Prediction and Description}
With the rapid development of large language models (LLMs), their application in the transportation domain has expanded considerably. Recent LLMs such as BERT \cite{bert}, GPT-2 \cite{gpt2}, GPT-3~\cite{gpt3}, GPT-4~\cite{gpt4}, DeepSeek~\cite{deepseek}, and LLaMA~\cite{llama} have demonstrated impressive capabilities in understanding and generating natural language, enabling new possibilities for traffic-related tasks. These models have been applied to various domains, including road network analysis, route planning, and traffic report generation, where their strong generalization and reasoning abilities have shown substantial benefits.

Several pioneering works have directly explored the synergy between LLMs and traffic prediction. For example, TrafficBERT~\cite{TrafficBERT} introduced a pre-trained language model tailored for traffic data, improving prediction accuracy by leveraging textual reports. Text-enhanced spatiotemporal graph neural networks~\cite{tesg} further demonstrate how integrating textual information can significantly benefit the forecasting of abnormal traffic events, inspiring the use of LLMs to more effectively exploit unstructured text. More recently, models such as STLLM~\cite{stllm} and TPLLM~\cite{tpllm} have integrated LLMs with spatiotemporal modeling architectures, achieving state-of-the-art results in multi-modal traffic forecasting. These approaches highlight the dual value of LLMs: enhancing numerical prediction by incorporating unstructured text, and enabling automated, context-aware generation of traffic condition descriptions—particularly for abnormal or non-recurrent events. The text generation capabilities of LLMs present new opportunities for interpreting and communicating complex traffic scenarios. By generating detailed and contextually relevant natural language descriptions alongside numerical forecasts, LLM-based systems can provide more transparent and actionable information to both transportation professionals and the general public. 

Despite these advances, there remains significant room for developing unified frameworks that fully leverage the strengths of LLMs across both predictive and descriptive modalities, especially in handling abnormal traffic events and bridging the gap between quantitative analysis and human-centric explanation.

\section{Methodology}
This section presents the architecture and components of our CrossTrafficLLM framework, a novel multi-modal approach that bridges traffic forecasting and textual description generation. Our model architecture is depicted in Fig. 2. The methodology of our proposed framework consists of two main components: (1)Traffic Prediction, (2) Text Generation

\begin{figure*}[!t]
\centering
\includegraphics[width=6in]{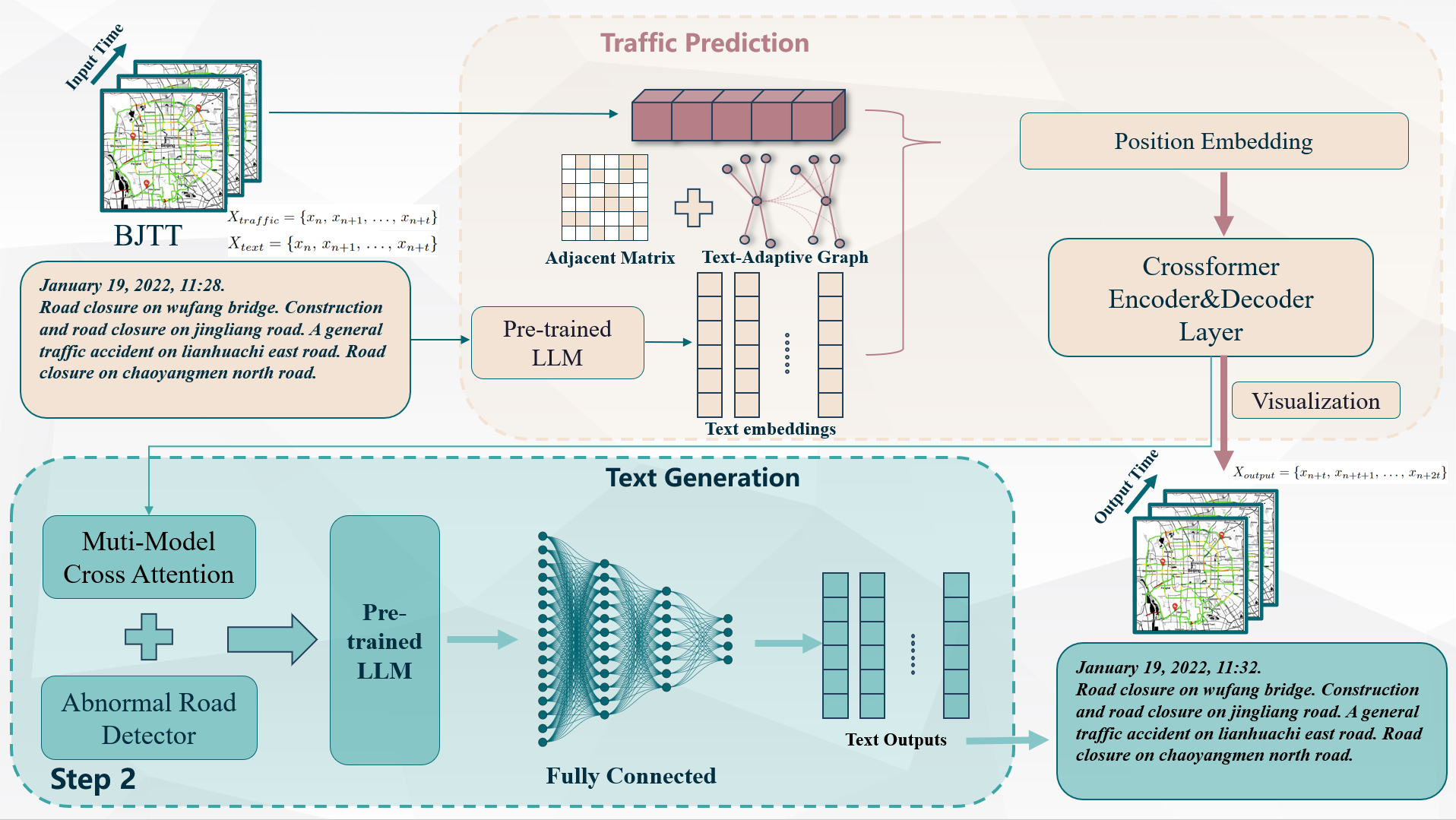}
\caption{Method overview. The proposed model comprises two key components: Traffic Prediction and Text Generation. The Traffic Prediction component integrates a Crossformer architecture, Large Language Model (LLM), and adaptive graph convolution. Textual features extracted by the LLM guide the adaptive matrix in processing traffic data before feeding into the Crossformer for prediction. The Text Generation component processes traffic data through Cross-attention and an Abnormal Road Detector layer before fine-tuning a large language model to generate descriptive text output.}
\label{fig_2}
\end{figure*}

\subsection{Problem Formulation}

We address the bidirectional relationship between spatiotemporal traffic data and contextual textual information. Formally, we define the problem as follows:

Given historical traffic data from time step $n$ to $n+t$ represented as $\mathbf{X} \in \mathbb{R}^{B \times t \times N \times C}$ (where $B$ is batch size, $t$ is sequence length, $N$ is the number of road nodes, and $C$ is the number of traffic features per node), a road network adjacency matrix $\mathbf{A} \in \mathbb{R}^{N \times N}$, and corresponding historical textual descriptions $\mathbf{T}_{hist}$ from time step $n$ to $n+t$, our objective is to:

\begin{enumerate}
\item Predict future traffic conditions $\hat{\mathbf{Y}} \in \mathbb{R}^{B \times t \times N \times C}$ for time steps $n+t$ to $n+2t$.
\item Generate contextually relevant textual descriptions $\hat{\mathbf{T}}_{gen}$ for time steps $n+t$ to $n+2t$ based on the predicted traffic conditions.
\end{enumerate}

This bidirectional framework leverages historical data from both modalities to generate predictions that maintain consistency between numerical traffic forecasts and their textual representations.

\subsection{Traffic Prediction}

The traffic prediction module in our framework enhances the Crossformer architecture with a novel text-guided adaptive graph convolutional network (GCN), enabling the integration of textual context into the spatiotemporal traffic forecasting process.

\subsubsection{TextCrossformer Architecture}

Our TextCrossformer extends the original Crossformer by incorporating textual information as auxiliary input, thereby improving the model's prediction capability. The model segments historical traffic data into fixed-length patches using a temporal patch embedding layer, followed by hierarchical encoding through multiple scale blocks. Each scale block contains a two-stage attention mechanism: temporal multi-head self-attention and feature-wise self-attention, which together capture complex dependencies across time and network nodes.

For future traffic prediction, the decoder applies cross-attention to integrate encoded historical information, and outputs are projected through a flattening head to generate the final forecasts across all nodes and time steps.

\subsubsection{Text Encoding and Feature Integration}

A core innovation is the integration of textual information into the traffic prediction process. We employ a DeepSeek-based encoder to process historical textual descriptions, extracting semantic features that supplement traffic data. Specifically, given a sequence of historical text tokens $\mathbf{T}_{\text{hist}} \in \mathbb{R}^{B \times S \times L}$, the text encoder produces:

\begin{equation}
    \mathbf{T}_{\text{embed}} = \text{TextEncoder}(\mathbf{T}_{\text{hist}})
\end{equation}
where $\mathbf{T}_{\text{embed}} \in \mathbb{R}^{B \times S \times D}$ is the sequence of semantic text embeddings, $B$ is the batch size, $S$ is the sequence length, $L$ is the token length, and $D$ is the embedding dimension. The embeddings are normalized and projected to match the traffic feature space.

\subsubsection{Text-Guided Adaptive Graph Convolutional Network}

This module integrates textual semantics into traffic prediction through a conditional feature modulation mechanism. Instead of simply concatenating modalities, we use text to dynamically recalibrate traffic features before they enter the graph network. The process consists of three stages: encoding, alignment, and modulation.

\paragraph{Location-Aware Text Encoding}
Standard embeddings often fail to capture domain-specific traffic phrases. We employ a parallel encoder to extract multi-scale semantic features. A 1D convolution layer detects local phrases (e.g., ``traffic jam''), while a depthwise separable convolution captures global syntactic dependencies. These features are fused to generate the text representation sequence $\mathbf{H}_{\text{text}}$.

\paragraph{Sparse Cross-Modal Alignment}
To align unstructured text with structured traffic time series, we use a \textbf{Sparse Interaction Mechanism}. We calculate the relevance between traffic features $\mathbf{H}_{\text{traffic}}$ and text embeddings $\mathbf{H}_{\text{text}}$. A Top-$K$ selection is applied to filter out noise (irrelevant words), ensuring only critical semantic tokens contribute to the context. The aligned textual context $\mathbf{C}_{\text{text}}$ is computed as:
\begin{equation}
    \mathbf{C}_{\text{text}} = \text{Softmax}(\text{TopK}(\mathbf{H}_{\text{traffic}} \mathbf{H}_{\text{text}}^T)) \cdot \mathbf{H}_{\text{text}}
\end{equation}
This operation maps specific text descriptions to their corresponding traffic time steps.

\paragraph{Text-Conditional Feature Modulation}
We employ a FiLM-like mechanism to inject the aligned context into traffic representations. The context $\mathbf{C}_{\text{text}}$ is projected to generate dynamic scaling ($\boldsymbol{\alpha}$) and shifting ($\boldsymbol{\beta}$) parameters:
\begin{equation}
    \boldsymbol{\alpha} = \sigma(\mathbf{W}_{\alpha} \mathbf{C}_{\text{text}}), \quad \boldsymbol{\beta} = \tanh(\mathbf{W}_{\beta} \mathbf{C}_{\text{text}})
\end{equation}
These parameters modulate the original traffic features:
\begin{equation}
    \mathbf{X}_{\text{guided}} = \boldsymbol{\alpha} \odot \mathbf{H}_{\text{traffic}} + \eta \cdot \boldsymbol{\beta}
\end{equation}
where $\odot$ denotes element-wise multiplication and $\eta$ is a scaling factor. Through this, text serves as a gate that amplifies or suppresses specific traffic signals (e.g., suppressing flow values during a ``road closure'') based on semantic meaning.

\paragraph{Adaptive Graph Convolution}
The modulated features $\mathbf{X}_{\text{guided}}$ then serve as input to the GCN. We construct an adaptive adjacency matrix $\mathbf{A}_{\text{adp}}$ using learnable node embeddings $\mathbf{E}$ to capture latent spatial dependencies:
\begin{equation}
    \mathbf{A}_{\text{adp}} = \text{Softmax}(\text{ReLU}(\mathbf{E}\mathbf{E}^T))
\end{equation}
The final spatial aggregation is performed as:
\begin{equation}
    \mathbf{H}_{\text{out}} = \text{GCNBlock}(\mathbf{X}_{\text{guided}}, \mathbf{A}_{\text{adp}})
\end{equation}

\paragraph{Variable Definitions}
\begin{itemize}
    \item $\mathbf{H}_{\text{text}}$: Encoded text feature sequence, shape $[B, L, D]$
    \item $\mathbf{H}_{\text{traffic}}$: Input traffic feature sequence, shape $[B, S, D]$
    \item $\mathbf{C}_{\text{text}}$: Aligned textual context matrix, shape $[B, S, D]$
    \item $\boldsymbol{\alpha}, \boldsymbol{\beta}$: Modulation parameters (scaling and shifting), shape $[B, S, D]$
    \item $\mathbf{X}_{\text{guided}}$: Text-modulated traffic features, shape $[B, S, D]$
    \item $\mathbf{E}$: Learnable node embedding dictionary, shape $[N, d]$
    \item $\mathbf{A}_{\text{adp}}$: Adaptive adjacency matrix, shape $[N, N]$
    \item $\mathbf{H}_{\text{out}}$: Output features after GCN processing
    \item $\eta$: Scaling factor for the bias term (set to 0.1)
\end{itemize}

\subsection{Text Generation}

The text generation component leverages predicted traffic conditions to generate contextually relevant descriptions, supported by several novel mechanisms.

\subsubsection{Road Importance Detection}

We introduce a road importance detection mechanism that assigns importance scores to each road segment based on traffic features (e.g., congestion, speed):

\begin{equation}
    \mathbf{I}_{\text{road}} = \sigma\left(\text{RoadDetector}\left(\mathbf{X}_{\text{traffic}}\right)\right)
\end{equation}
where $\mathbf{I}_{\text{road}} \in \mathbb{R}^{B \times T \times N \times 1}$ denotes the importance scores, $T$ is the number of time steps, and $\sigma(\cdot)$ is the sigmoid function. The model selects the top $30\%$ of roads with the highest importance for efficient and focused text generation, creating an importance-weighted summary of critical traffic conditions.

\subsubsection{Road-Aware Cross-Attention}

To effectively fuse traffic features into text generation, we propose a road-aware cross-attention mechanism. Text features ($\mathbf{H}_{\text{text}}$) act as queries, while traffic features ($\mathbf{H}_{\text{modulated}}$) serve as keys and values, modulated by road importance and controlled by a dynamic gating mechanism:

\begin{equation}
    \mathbf{H}_{\text{fusion}} = \mathbf{H}_{\text{text}} + \mathbf{G} \odot \mathbf{H}_{\text{modulated}}
\end{equation}
where $\mathbf{G}$ is a learned gate of the same shape as $\mathbf{H}_{\text{text}}$, and $\odot$ denotes element-wise multiplication. This ensures that more relevant traffic information more strongly influences the generated description.

\subsubsection{Road-Text Memory Module}

To maintain semantic and linguistic consistency, we employ a road-text memory module based on multi-head attention, which learns long-term associations between traffic patterns and their textual descriptions. This enables the model to recall how specific traffic conditions were previously described, improving coherence and factual consistency.

\subsubsection{Traffic-Enhanced Text Generation}

All aforementioned components are integrated into the DeepSeek-based language model, which generates the final textual output. The token generation probability at time $t$ is given by:

\begin{equation}
    P(w_t | \mathbf{w}_{<t}, \mathbf{X}_{\text{traffic}}) = \text{softmax}\left(\mathbf{LMHead}\left(\mathbf{H}_{\text{fusion}}\right)\right)
\end{equation}
where $w_t$ is the token at time step $t$, $\mathbf{w}_{<t}$ are the previously generated tokens, and $\mathbf{LMHead}$ is the language model projection head.

\paragraph{Variable Definitions}
\begin{itemize}
    \item $\mathbf{X}_{\text{traffic}}$: Predicted traffic features, shape $[B, T, N, F]$
    \item $\mathbf{I}_{\text{road}}$: Road importance scores, shape $[B, T, N, 1]$
    \item $\mathbf{H}_{\text{text}}$: Text feature representations, shape $[B, T, L, D]$
    \item $\mathbf{H}_{\text{modulated}}$: Traffic features modulated by importance, shape $[B, T, L, D]$
    \item $\mathbf{G}$: Dynamic gate, shape $[B, T, L, D]$
    \item $\mathbf{H}_{\text{fusion}}$: Fused features for text generation, shape $[B, T, L, D]$
    \item $\mathbf{LMHead}$: The final projection (language model) head
    \item $w_t$: The token generated at time step $t$
\end{itemize}

\paragraph{Summary}  
Our framework tightly couples spatiotemporal traffic prediction and text generation, allowing textual context to influence prediction and traffic forecasts to guide generation. This results in highly accurate, context-aware, and interpretable traffic situation descriptions.

\section{Experiments}

In this section, we evaluate the performance of our CrossTrafficLLM framework through massive experiments on a large-scale multimodal traffic dataset. We first introduce the dataset and experimental settings, followed by comparisons with state-of-the-art methods in both traffic prediction and text generation tasks.

\subsection{Experimental Setup}

\subsubsection{Dataset}

We conduct experiments on the Beijing Traffic-Text (BjTT) dataset, a large-scale multimodal dataset introduced by Zhang et al. \cite{bjtt}. This huge dataset captures both traffic dynamics and textual descriptions across Beijing's urban road network. The dataset encompasses 1,260 road segments within Beijing's Fifth Ring Road, providing traffic speed measurements recorded at 4-minute intervals over 32,400 time steps, spanning approximately 90 days. 

The dataset includes three key components:
\begin{itemize}
    \item \textbf{Traffic Speed Data}: Time-series measurements of average vehicle speeds on each road segment, captured every 4 minutes.
    \item \textbf{Road Network Topology}: A road adjacency matrix $\mathbf{A} \in \mathbb{R}^{1260 \times 1260}$ that represents the connectivity between road segments in Beijing's urban transportation network.
    \item \textbf{Textual Descriptions}: Corresponding natural language descriptions of traffic conditions, with particular emphasis on non-routine events such as accidents, construction, road closures, and congestion.
\end{itemize}

The textual descriptions provide contextual information about traffic conditions, typically focusing on anomalous events. Each description includes location references and event details. 

For example: "January 01, 2022, 00:00. Road closure on Wufang Bridge. Construction and road closure on Jingliang Road. A general traffic accident on West Fourth Ring North Road. Construction and road closure on South Third Ring West Road..."

This multimodal dataset is particularly suitable for our research as it allows us to explore the bidirectional relationship between traffic patterns and their textual descriptions in a real-world urban setting.

\subsubsection{Baseline Methods}

To comprehensively evaluate our proposed framework, we compare it against state-of-the-art methods in both traffic prediction and text generation domains.

\textbf{Traffic Prediction Baselines:}
\begin{itemize}
    \item \textbf{STGCN} \cite{stgcn}: A pioneering work that combines graph convolutions and 1D convolutions to model spatial and temporal dependencies in traffic networks.
    \item \textbf{GWN} \cite{gwn}: Graph WaveNet, which uses dilated causal convolutions and adaptive adjacency matrices to capture multi-scale traffic patterns.
    \item \textbf{STSGCN} \cite{stsgcn}: Spatial-Temporal Synchronous Graph Convolutional Networks, which models local spatial-temporal correlations synchronously.
    \item \textbf{ASTGCN} \cite{astgcn}: Attention-based Spatial-Temporal Graph Convolutional Networks, which incorporates attention mechanisms to capture dynamic spatial-temporal correlations.
    \item \textbf{MTGNN} \cite{mtgnn}: Multivariate Time Series Graph Neural Networks, which learns graph structures adaptively and employs a mix-hop propagation layer.
    \item \textbf{STG-NCDE} \cite{stg-ncde}: Spatial-Temporal Graph Neural Controlled Differential Equations, which integrates neural ODEs with graph neural networks for continuous-time modeling.
    \item \textbf{Crossformer} \cite{crossformer}: A novel architecture that introduces a two-stage attention mechanism with segment-wise self-attention and cross-time attention to effectively capture both local and global temporal dependencies while maintaining computational efficiency for long sequence time series forecasting.
    \item \textbf{Informer} \cite{informer}: An efficient Transformer-based model designed for long sequence time-series forecasting, which uses a ProbSparse self-attention mechanism to reduce computational complexity and enhance the modeling of long-range dependencies.
    \item \textbf{iTransformer} \cite{itransformer}: An inverted Transformer architecture for time series forecasting, which applies attention across the variate (variable) dimension instead of the temporal dimension, enabling more effective modeling of multivariate correlations and better scalability to longer look-back windows without modifying the basic Transformer components.
    \item \textbf{TimeXer} \cite{timexer}: A Transformer-based framework tailored for time series forecasting with exogenous variables, which employs differentiated embedding strategies and cross-attention mechanisms to effectively integrate external information and boost prediction performance.
\end{itemize}

We selected these baselines because they represent the evolution of spatial-temporal graph neural networks for traffic prediction. They cover various approaches from basic GCN-based models (STGCN) to more sophisticated techniques that incorporate adaptive graph structures (GWN, MTGNN) and continuous-time dynamics (STG-NCDE).

\textbf{Text Generation Baselines:}
\begin{itemize}
    \item \textbf{CLIPCap} \cite{clip}: A simple method that uses CLIP image encodings as a prefix for a fine-tuned GPT-2 language model to generate captions efficiently with minimal training.
    \item \textbf{BLIP} \cite{blip}: Bootstrapping Language-Image Pre-training, a vision-language framework that improves noisy web data utilization by bootstrapping captions, filtering poor-quality data to achieve state-of-the-art performance across multiple vision-language tasks.
    \item \textbf{Transformer} \cite{attention}: The model selects the most probable next word at each step until text generation is complete.
    \item \textbf{DiffCap} \cite{diff}: A novel approach that transforms discrete tokens and applies continuous diffusion models to generate diverse captions, achieving competitive results with a simpler structure compared to existing non-autoregressive models.
    \item \textbf{DDCap} \cite{ddcap}: A diffusion-based image captioning model offering greater flexibility than traditional auto-regressive methods, using techniques like best-first inference and text length prediction to tackle text decoding challenges.
\end{itemize}

These text generation baselines were selected for their ability to generate descriptive text from non-textual inputs. While originally designed for image captioning, these methods have been adapted in our experiments to generate traffic descriptions from numerical traffic data.

\subsubsection{Evaluation Metrics}

We employ standard metrics in both traffic prediction and text generation domains to evaluate our model's performance.

\textbf{Traffic Prediction Metrics:}
\begin{itemize}
    \item \textbf{Mean Absolute Error (MAE)}:
    \begin{equation}
        \text{MAE} = \frac{1}{n} \sum_{i=1}^{n} |y_i - \hat{y}_i|
    \end{equation}
    where $y_i$ and $\hat{y}_i$ are the ground truth and predicted values, respectively, and $n$ is the number of samples.
    
    \item \textbf{Root Mean Square Error (RMSE)}:
    \begin{equation}
        \text{RMSE} = \sqrt{\frac{1}{n} \sum_{i=1}^{n} (y_i - \hat{y}_i)^2}
    \end{equation}
    RMSE penalizes larger errors more severely than MAE, providing insight into prediction stability.
\end{itemize}

\textbf{Text Generation Metrics:}
\begin{itemize}
    \item \textbf{BLEU-4} \cite{bleu}: Evaluates the precision of generated text by comparing n-gram matches between the generated text and reference texts:
    \begin{equation}
        \text{BLEU-4} = BP \cdot \exp\left(\sum_{n=1}^{4} w_n \log p_n\right)
    \end{equation}
    where $p_n$ is the modified n-gram precision, $w_n$ is the weight for each n-gram precision, and BP is the brevity penalty.
    
    \item \textbf{METEOR} \cite{meteor}: Evaluates the harmonic mean of precision and recall, with consideration for stemming and synonymy:
    \begin{equation}
        \text{METEOR} = (1 - \gamma \cdot \text{frag}^{\beta}) \cdot \frac{P \cdot R}{\alpha \cdot P + (1 - \alpha) \cdot R}
    \end{equation}
    where $P$ is precision, $R$ is recall, $\text{frag}$ is fragmentation penalty, and $\alpha$, $\beta$, $\gamma$ are parameters.
    
    \item \textbf{ROUGE-L} \cite{rouge}: Measures the longest common subsequence between generated and reference texts:
    \begin{equation}
        \text{ROUGE-L} = \frac{(1 + \beta^2) \cdot LCS_P \cdot LCS_R}{\beta^2 \cdot LCS_P + LCS_R}
    \end{equation}
    where $LCS_P$ and $LCS_R$ are precision and recall based on the longest common subsequence.
\end{itemize}

These metrics comprehensively evaluate both the accuracy of traffic predictions and the quality of generated textual descriptions.

\subsubsection{Implementation Details}

We implement our CrossTrafficLLM framework using PyTorch. For the language model component, we use the DeepSeek-R1-Distill-1.5B model, which has a hidden dimension of 1,536. To efficiently fine-tune this large language model while preserving its pre-trained knowledge, we employ Low-Rank Adaptation (LoRA) with rank $r=12$, scaling factor $\alpha=24$, and dropout rate of 0.1.

The model is trained end-to-end with the Adam optimizer using a learning rate of $5 \times 10^{-5}$ and a batch size of 4. We train the model for 50 epochs with a linear learning rate warm-up for the first 5 epochs followed by a cosine annealing schedule. For traffic prediction, we use a sequence length of 12 time steps (48 minutes) to predict the next 12 time steps.

For cross-validation, we split the dataset into training (80\%), and test (20\%) sets, ensuring temporal continuity. All experiments are conducted on a single NVIDIA GeForce RTX 4090 GPU with 24GB memory.

\subsection{Results and Analysis}

\subsubsection{Traffic Prediction Performance}

Table~\ref{tab:model_comparison} presents the comprehensive performance comparison of different traffic prediction models on the BjTT dataset across multiple prediction horizons. To thoroughly evaluate temporal generalization capabilities, we tested each model on three distinct forecasting intervals: 20 minutes (T=5), 40 minutes (T=10), and 60 minutes (T=15), with each time step representing a 4-minute interval.

\begin{table*}[!t]
\caption{Performance Comparison of Different Traffic Prediction Models. T=5(20min), 10(40min), 15(60min). The best performance is in bold.}
\label{tab:model_comparison}
\centering
\renewcommand{\arraystretch}{1.2}
\begin{tabular}{c|>{\centering\arraybackslash}p{1.5cm}>{\centering\arraybackslash}p{1.5cm}|>{\centering\arraybackslash}p{1.5cm}>{\centering\arraybackslash}p{1.5cm}|>{\centering\arraybackslash}p{1.5cm}>{\centering\arraybackslash}p{1.5cm}} 
\hline
\multirow{2}{*}{Model} & \multicolumn{2}{c|}{T=5 , 20min} & \multicolumn{2}{c|}{T=10 , 40min} & \multicolumn{2}{c}{T=15 , 60min} \\ 
\hhline{~------}
                     & MAE & RMSE & MAE & RMSE & MAE & RMSE \\ 
\hline
STGCN \cite{stgcn}                & 3.56 & 5.45 & 3.85 & 5.93 & 4.00 & 6.19 \\
GWN \cite{gwn}                  & 3.61 & 5.60 & 3.92 & 6.19 & 4.16 & 6.61 \\
ASTGCN \cite{astgcn}               & 3.79 & 5.87 & 4.15 & 6.54 & 4.44 & 7.02 \\
STSGCN \cite{stsgcn}               & 3.66 & 5.74 & 4.01 & 6.22 & 4.15 & 6.63 \\
MTGNN \cite{mtgnn}                & 3.60 & 5.60 & 3.74 & 5.91 & 3.89 & 6.19 \\
STG-NCDE \cite{stg-ncde}             & 3.56 & 5.53 & 3.83 & 6.09 & 4.05 & 6.52 \\
Informer \cite{informer}             & 3.55 & 5.36 & 3.84 & 5.85 & 4.02 & 6.13 \\
iTransformer \cite{itransformer}             & 3.83 & 5.91 & 3.84 & 5.93 & 3.86 & 5.99 \\
TimeXer \cite{timexer}             & 3.43 & 5.21 & 3.61 & 5.48 & 3.74 & 5.63 \\
Crossformer \cite{crossformer}          & 3.39 & 5.03 & 3.52 & 5.31 & 3.79 & 5.51 \\
CrossTrafficLLM(no Text) & 3.31 & 4.92 & 3.45 & 5.19 & 3.64 & 5.46 \\ 
\hline
CrossTrafficLLM       & \textbf{3.25} & \textbf{4.86} & \textbf{3.37} & \textbf{5.11} & \textbf{3.55} & \textbf{5.36} \\
\hline
\end{tabular}
\end{table*}

From the results, we observe that our proposed CrossTrafficLLM framework consistently achieves the best performance across all prediction horizons, demonstrating its superior forecasting capabilities for both short-term and longer-term traffic predictions. Several critical insights emerge from this multi-horizon analysis:

\textbf{Consistent performance advantage:} CrossTrafficLLM maintains its superiority over all baseline models across all time horizons, with the performance advantage being particularly pronounced for longer prediction intervals. At T=15 (60 minutes), our model achieves an MAE of 3.55 and RMSE of 5.36, representing significant improvements over traditional spatial-temporal graph neural network models: 11.3\% MAE reduction compared to STGCN, 14.7\% compared to GWN, and 20.0\% compared to ASTGCN.

\textbf{Transformer-based models:} Among the baselines, recent transformer-based models such as Informer, iTransformer, TimeXer, and Crossformer show strong forecasting ability, especially for short- and mid-term horizons. Notably, TimeXer and Crossformer achieve lower MAE and RMSE than most graph-based models, confirming the effectiveness of advanced attention mechanisms in capturing long-range spatiotemporal dependencies. For example, TimeXer achieves an MAE of 3.43 for T=5, while Crossformer achieves 3.39. However, even these models are consistently outperformed by our CrossTrafficLLM, demonstrating the added value of integrating textual information and adaptive graph learning.

\textbf{Temporal degradation resilience:} While all models exhibit some performance degradation as the prediction horizon extends (a common challenge in traffic forecasting), CrossTrafficLLM demonstrates remarkable resilience. The performance decline from T=5 to T=15 is notably smaller for our model (9.2\% increase in MAE) compared to GWN (15.2\%), ASTGCN (17.2\%), and STSGCN (13.4\%). This enhanced temporal stability can be attributed to our text-guided approach, which incorporates contextual information that remains relevant over longer horizons.

\textbf{Short-term precision:} At T=5 (20 minutes), CrossTrafficLLM achieves an MAE of 3.25 and RMSE of 4.86, outperforming the next best model (CrossTrafficLLM without text) by 1.8\% in MAE and 1.2\% in RMSE. Compared to the best transformer baselines (e.g., Crossformer and TimeXer), our model achieves 4.1\% and 5.2\% lower MAE, respectively, underlining the contribution of textual information even when traffic patterns are more predictable.

\textbf{Mid-term forecasting effectiveness:} The performance at T=10 (40 minutes) shows that CrossTrafficLLM maintains its advantage in the critical mid-term prediction range, which is particularly valuable for practical traffic management applications such as signal timing adjustments and congestion mitigation strategies. In this range, our model achieves an MAE of 3.37 and RMSE of 5.11, outperforming the pure Crossformer architecture by 4.3\% in MAE and the TimeXer by 6.7\%.

The consistent performance gains across all time horizons can be attributed to several key architectural innovations:

\textbf{Two-stage attention mechanism:} The underlying CrossFormer architecture, with its two-stage attention mechanism designed to capture temporal and spatial correlations separately, provides a strong foundation. This hierarchical design effectively captures multi-scale temporal dependencies in traffic data, as evidenced by Crossformer's strong performance across all prediction horizons (MAE: 3.39/3.52/3.79 for T=5/10/15).
    
\textbf{Adaptive matrix integration:} The integration of adaptive adjacency matrices further enhances prediction accuracy, as demonstrated by the CrossTrafficLLM(no Text) variant. This component enables more effective modeling of dynamic spatial relationships within the traffic network, where connectivity patterns vary based on evolving traffic conditions. The improvements are consistent across all time horizons (2.4\% for T=5, 2.0\% for T=10, and 4.0\% for T=15 in MAE reduction compared to Crossformer).
    
\textbf{Text-guided prediction:} Most significantly, incorporating textual information through our text-guided GCN yields additional performance gains across all prediction horizons. The improvement from CrossTrafficLLM(no Text) to the full CrossTrafficLLM model ranges from 1.8\% MAE reduction for T=5 to 2.5\% for T=15, with RMSE reductions of 1.2\% to 1.8\% respectively. This increasing benefit over longer horizons suggests that textual information becomes increasingly valuable for predictions further into the future, where numerical patterns alone may be insufficient.

These multi-horizon results significantly strengthen our findings, confirming that textual information provides complementary insights to numerical traffic data across various prediction timescales. The contribution of textual context appears particularly pronounced for longer prediction horizons, where traditional models struggle with increasing uncertainty. The textual descriptions often contain information about planned events, ongoing construction, or recurring patterns that may not be fully captured by historical traffic data alone, enabling more accurate forecasting of future traffic states.

Moreover, the consistent performance across different horizons demonstrates the robustness of our approach and its potential applicability in practical traffic management systems, where prediction accuracy at multiple time scales is essential for effective decision-making. The ability to maintain relatively stable performance as the prediction horizon extends represents a significant advancement in traffic forecasting technology.

\subsubsection{Ablation Analysis on Prediction Component}
We analyze the contribution of the textual module to the forecasting task. As shown in Table~\ref{tab:model_comparison}, we compare the full \textbf{CrossTrafficLLM} against \textbf{CrossTrafficLLM(no Text)}, which removes the text-guided adaptive GCN branch and relies solely on numerical data.
The results indicate a clear performance drop when textual semantics are excluded. For instance, at T=15, the MAE increases from 3.55 to 3.64 (+2.5\%), and RMSE increases from 5.36 to 5.46 (+1.9\%). This degradation confirms that the unstructured text is not merely redundant noise but contains high-level semantic signals (e.g., the severity of an accident) that the numerical series alone cannot fully capture. This explicitly validates our methodological hypothesis: aligning semantic context with graph structure fundamentally improves the model's ability to anticipate complex traffic dynamics.

\subsubsection{Text Generation Performance}

Table~\ref{tab:caption_comparison} presents the performance comparison of different text generation models. Our CrossTrafficLLM framework significantly outperforms all baseline methods across all evaluation metrics.

\begin{table}[!t]
\caption{Performance Comparison of Different Caption Generation Models.As the generated texts at nearby time steps are often highly similar, we report the averaged results across multiple time steps for a more representative assessment. The best performance is in bold.}
\label{tab:caption_comparison}
\centering
\renewcommand{\arraystretch}{1.2}
\begin{tabular}{l|ccc}
\hline
\textbf{Model} & \textbf{BLEU-4} & \textbf{METEOR} & \textbf{ROUGE-L} \\
\hline
CLIPCap \cite{clip} & - & 3.42 & 23.70 \\
BLIP \cite{blip} & 1.29 & 14.36 & 50.95 \\
Transformer \cite{attention} & 6.34 & 20.22 & 60.55 \\
DiffCap \cite{diff} & 44.36 & 51.28 & 65.83 \\
DDCap \cite{ddcap} & 56.29 & 63.85 & 84.46 \\
\hline
CrossTrafficLLM & \textbf{71.58} & \textbf{72.56} & \textbf{89.64} \\
\hline
\end{tabular}
\end{table}

The results reveal several important insights:

\begin{enumerate}
    \item \textbf{Superior performance:} Our CrossTrafficLLM framework achieves remarkable improvements across all metrics, with a BLEU-4 score of 71.58, METEOR score of 72.56, and ROUGE-L score of 89.64. These results represent substantial gains over the next best model (DDCap), with improvements of 27.2\% in BLEU-4, 13.6\% in METEOR, and 6.1\% in ROUGE-L.
    
    \item \textbf{Large language model advantage:} The foundation of our text generation component on a pre-trained large language model (DeepSeek) provides a significant advantage in generating coherent and contextually relevant textual descriptions. This is evident in the substantial performance gap between our approach and traditional encoder-decoder models like the Transformer baseline.
    
    \item \textbf{Traffic-aware generation:} The integration of traffic-specific features through our road importance detection and cross-attention mechanisms enables the generation of highly accurate and detailed traffic descriptions. This domain-specific adaptation is crucial for capturing the specialized vocabulary and structure of traffic reports.
    
    \item \textbf{Vision-based limitations:} Models originally designed for image captioning, such as CLIPCap and BLIP, demonstrate limited performance in this domain-specific task. CLIPCap fails to produce meaningful BLEU-4 scores, while BLIP achieves only 1.29, indicating that visual-textual pre-training does not transfer effectively to traffic-textual tasks without substantial adaptation.
\end{enumerate}

To provide a more intuitive comparison of text generation quality, Fig. 3 presents examples of text generated by CrossTrafficLLM and baseline models. The red portions highlight content that does not match the ground truth. Our model produces the most complete and accurate textual descriptions, with only minor discrepancies in event classification. In contrast, diffusion-based models like DiffCap and DDCap exhibit more significant errors, including incomplete sentences and missing critical information. BLIP and CLIPCap, while occasionally generating correct content, omit substantial portions of the relevant traffic information.

The superior performance of CrossTrafficLLM in text generation can be attributed to several factors: (1) the foundation in a powerful language model, (2) the road importance detection mechanism that focuses on critical traffic events, (3) the road-aware cross-attention that aligns traffic features with textual elements, and (4) the road-text memory module that maintains consistency between traffic conditions and their textual descriptions.

Importantly, our model demonstrates robust performance across varying traffic conditions, generating accurate reports for both routine traffic patterns and anomalous events. This versatility is essential for real-world deployment, where traffic reporting systems must handle both everyday congestion and unexpected incidents.

\begin{figure*}[!t]
\centering
\includegraphics[width=7.5in]{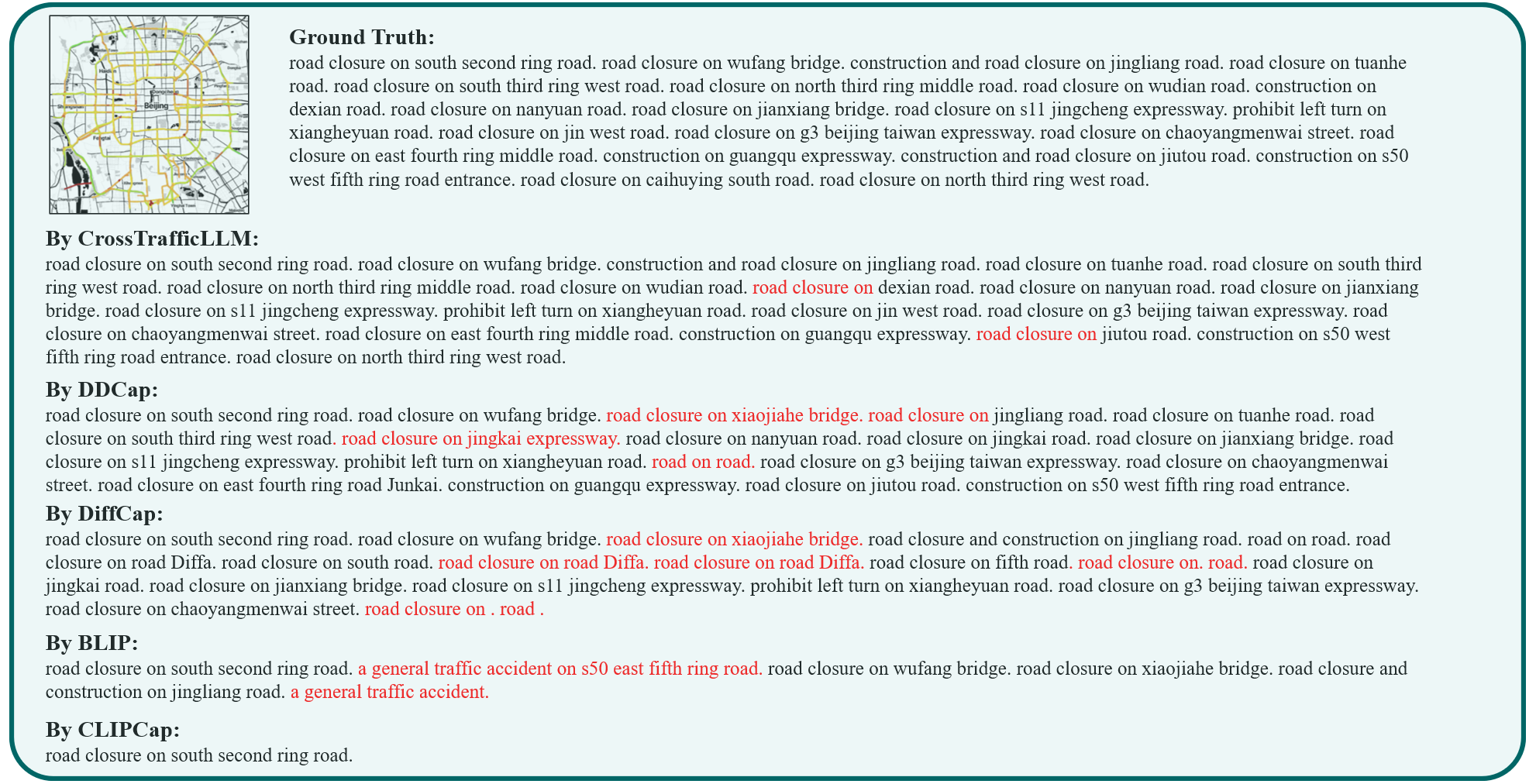}
\caption{An example of text generated by DDCap, DiffCap, BLIP, Transformer and CLIPCap. The red portions highlight content that does not match the ground truth. There is a high consistency between the text generated by our CrossTrafficLLM and the ground truth text, with only a a few event in this example deviating.}
\label{fig_3}
\end{figure*}

\subsection{Ablation Study}

To evaluate the contribution of each key component in our CrossTrafficLLM framework, we conduct a series of ablation experiments. Table~\ref{tab:ablation} presents the performance of different model variants, with components progressively added to demonstrate their individual and combined effects on text generation quality.

\begin{table*}[!t]
\caption{Ablation Study on Model Components. As the generated texts at nearby time steps are often highly similar, we report the averaged results across multiple time steps for a more representative assessment. The best performance is in bold.}
\label{tab:ablation}
\centering
\renewcommand{\arraystretch}{1.2}
\begin{tabular}{lccc|ccc}
\hline
\multicolumn{4}{c|}{\textbf{Model Components}} & \multicolumn{3}{c}{\textbf{Evaluation Metrics}} \\
\hline
GCN & Road Importance & Cross Attention & Text\_Traffic memory & BLEU-4 & ROUGE-L & METEOR \\
\hline
$-$            & $-$           & $-$          & $-$                & 66.69 & 79.57 & 67.10 \\
$\checkmark$   & $-$           & $-$          & $-$                & 67.38 & 82.36 & 66.40 \\
$\checkmark$   & $\checkmark$  & $-$          & $-$                & 68.21 & 83.28 & 68.24 \\
$\checkmark$   & $\checkmark$  & $\checkmark$ & $-$                & 71.28 & 88.95 & 71.28 \\
\hline
$\checkmark$   & $\checkmark$  & $\checkmark$ & $\checkmark$       & \textbf{71.58} & \textbf{89.64} & \textbf{72.56} \\
\hline
\end{tabular}
\end{table*}

\subsubsection{Impact of Individual Components}

Our ablation study reveals the following insights about each component:

\textbf{1. Graph Convolutional Network (GCN):} 
The introduction of GCN to process the spatial relationships between road segments provides a modest but meaningful improvement. Compared to the baseline model without any specialized components, incorporating GCN increases the BLEU-4 score from 66.69 to 67.38 (+1.0\%) and ROUGE-L from 79.57 to 82.36 (+3.5\%). This improvement demonstrates that capturing the topological structure of the road network enhances the model's ability to generate spatially coherent traffic descriptions. The GCN enables the model to understand which road segments are connected, allowing it to generate more accurate descriptions of how traffic conditions propagate through the network.

\textbf{2. Road Importance Detection:} 
Adding the road importance detection mechanism further improves performance across all metrics, with BLEU-4 increasing to 68.21 (+1.2\% relative to GCN only) and METEOR reaching 68.24 (+2.8\% relative to GCN only). This component allows the model to identify and prioritize critical road segments with significant traffic events, focusing computational resources and textual descriptions on the most relevant parts of the road network. By detecting abnormal or congested road segments, the model generates more precise and focused traffic reports that highlight meaningful events rather than routine conditions, which aligns with how human experts would report traffic situations.

\textbf{3. Cross-Attention Mechanism:} 
The road-aware cross-attention mechanism yields the most substantial single-component improvement, increasing BLEU-4 to 71.28 (+4.5\% relative to model with GCN and road importance) and ROUGE-L to 88.95 (+6.8\% relative to model with GCN and road importance). This significant improvement demonstrates the critical role of effectively aligning traffic features with textual elements. The cross-attention mechanism enables the model to selectively attend to relevant traffic features when generating each token in the output text, resulting in more accurate and contextually appropriate descriptions. This component acts as a bridge between numerical traffic representations and natural language generation, facilitating the translation of complex traffic patterns into coherent textual narratives.

\textbf{4. Text-Traffic Memory Module:} 
The addition of the text-traffic memory module yields further improvements across all metrics, with the full model achieving the best performance: BLEU-4 of 71.58, ROUGE-L of 89.64, and METEOR of 72.56. While the absolute improvement appears modest compared to the cross-attention component (+0.4\% for BLEU-4, +0.8\% for ROUGE-L, and +1.8\% for METEOR), this module plays a crucial role in maintaining consistency in the generated text over time. The memory module helps the model remember associations between specific traffic patterns and their textual descriptions, ensuring that similar traffic conditions receive similar linguistic treatment throughout the generated text. This consistency is particularly important for producing natural and human-like traffic reports that maintain coherent terminology and descriptive styles.

\subsubsection{Synergistic Effects}

The ablation study also reveals important synergistic effects between components. The combination of GCN and road importance detection provides a stronger foundation for traffic understanding than either component alone. Similarly, the cross-attention mechanism becomes more effective when operating on traffic representations enhanced by the GCN and prioritized by the road importance detector.

The full model with all four components demonstrates the most robust performance, indicating that each component contributes uniquely to the overall text generation capability. The progressive improvement with each added component confirms our architectural design choices and highlights the importance of a multi-faceted approach to traffic-aware text generation.

\subsubsection{Analysis of Component Contributions}

When analyzing the relative contributions of each component, we observe that:

1. The \textbf{cross-attention mechanism} provides the largest single performance boost (approximately 4.5\% in BLEU-4 and 6.8\% in ROUGE-L), highlighting the importance of effectively bridging traffic data and textual representations.

2. The \textbf{road importance detection} mechanism contributes notably to precision metrics (METEOR improves by 2.8\% relative to GCN only), suggesting that focusing on critical road segments helps generate more accurate descriptions of specific traffic events.

3. The \textbf{GCN component} enhances the structural understanding of the road network, with particularly strong improvements in ROUGE-L (+3.5\% compared to baseline), indicating better capture of longer text sequences that describe interconnected road segments.

4. The \textbf{text-traffic memory module} provides more modest but still meaningful improvements, with its largest impact on METEOR (+1.8\%), suggesting enhanced semantic alignment between generated and reference texts.

These findings confirm that our approach, incorporating spatial understanding, importance-based focusing, cross-modal alignment, and memory-based consistency, creates a robust framework for generating high-quality traffic descriptions. Each component addresses a different aspect of the challenging traffic-to-text task, collectively enabling our model to achieve state-of-the-art performance.

\subsection{Discussion}

While CrossTrafficLLM demonstrates strong performance in both traffic prediction and text generation, several limitations remain. First, the current framework primarily leverages structured traffic data, which may not fully capture the complex and dynamic nature of urban traffic, especially under the influence of external factors such as weather, social media reports, or special events. Integrating additional unstructured or multimodal information—such as real-time weather data, social media streams, images, and video—could further enhance the model’s understanding of abnormal or non-recurrent traffic patterns.

Second, in the present design, textual information is mainly utilized to guide the graph-based modeling of spatial dependencies. Future work could more deeply integrate textual features with temporal modeling in the transformer backbone, enabling richer cross-modal interactions and improved capture of evolving traffic situations.

Moreover, with the rapid advancement of large-scale pre-trained language models (e.g., GPT-4, Gemini 2.5), our framework could benefit from incorporating even more powerful models to boost the expressiveness and diversity of generated traffic reports. Additionally, constructing or leveraging knowledge graphs may help the model develop a more comprehensive understanding of complex urban traffic scenarios, further improving interpretability and generation quality.

Finally, due to current dataset constraints, our model’s real-time applicability and generalizability to other cities or data domains remain to be fully validated. Future research will focus on expanding data sources, optimizing real-time performance, and exploring deployment strategies in operational intelligent transportation systems, thereby broadening the practical scope and impact of multi-modal traffic analysis and generation.

\section{Conclusion}
\section{Conclusion}
In this work, we introduce CrossTrafficLLM, a framework developed with practical traffic management needs in mind. Our goal is to bring together spatiotemporal prediction and natural language generation in a way that reflects the social nature of traffic systems and the way human operators reason about them. While many existing models treat traffic anomalies as little more than numerical irregularities, our approach takes a different direction: through a Text-Guided Adaptive Graph Convolutional Network, the model links these anomalies to their semantic causes, allowing it to interpret unusual patterns in a more context-aware manner.

A central contribution of CrossTrafficLLM is its ability to produce coherent, meaningful descriptions of abnormal traffic states. Sudden deviations often confuse purely data-driven models, but our system can differentiate routine congestion from disruptions triggered by specific events such as accidvents or severe weather. These textual explanations offer traffic operators a clearer picture of what the model has detected and why the detection matters, ultimately improving situational awareness and making the underlying reasoning process less opaque.

Looking ahead, we plan to extend this generative framework in two directions. The first is to incorporate a broader range of real-time, unstructured data sources—such as social media feeds or emergency dispatch reports—to better reflect collective behavior during unusual situations. The second is to explore more interactive forms of human–AI collaboration, moving beyond one-shot descriptions toward a dialogue-based system that can answer causal questions or help evaluate potential interventions. By strengthening the model’s interpretability around abnormal conditions, we hope to contribute to more resilient and socially informed approaches to traffic governance.

\bibliographystyle{IEEEtran}
\bibliography{main}

\begin{IEEEbiography}[{\includegraphics[width=1in,height=1.25in,clip,keepaspectratio]{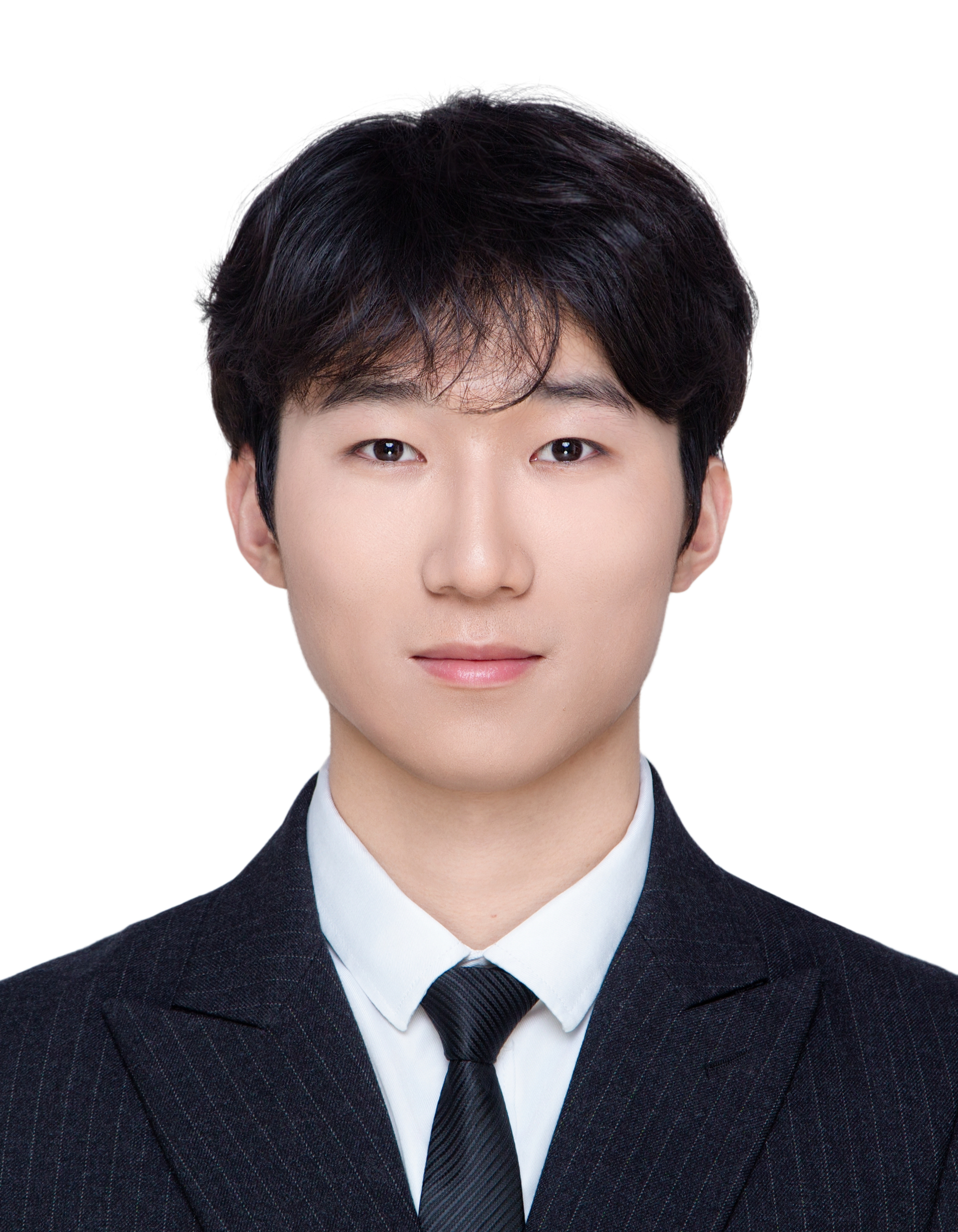}}]{Zeming Du} received the bachelor’s degree in computer science and technology from the Beijing University of Technology, China, in 2025, where he is currently pursuing the master’s degree in Artificial Intelligence at King's College London.
\end{IEEEbiography}

\begin{IEEEbiography}[{\includegraphics[width=1in,height=1.25in,clip,keepaspectratio]{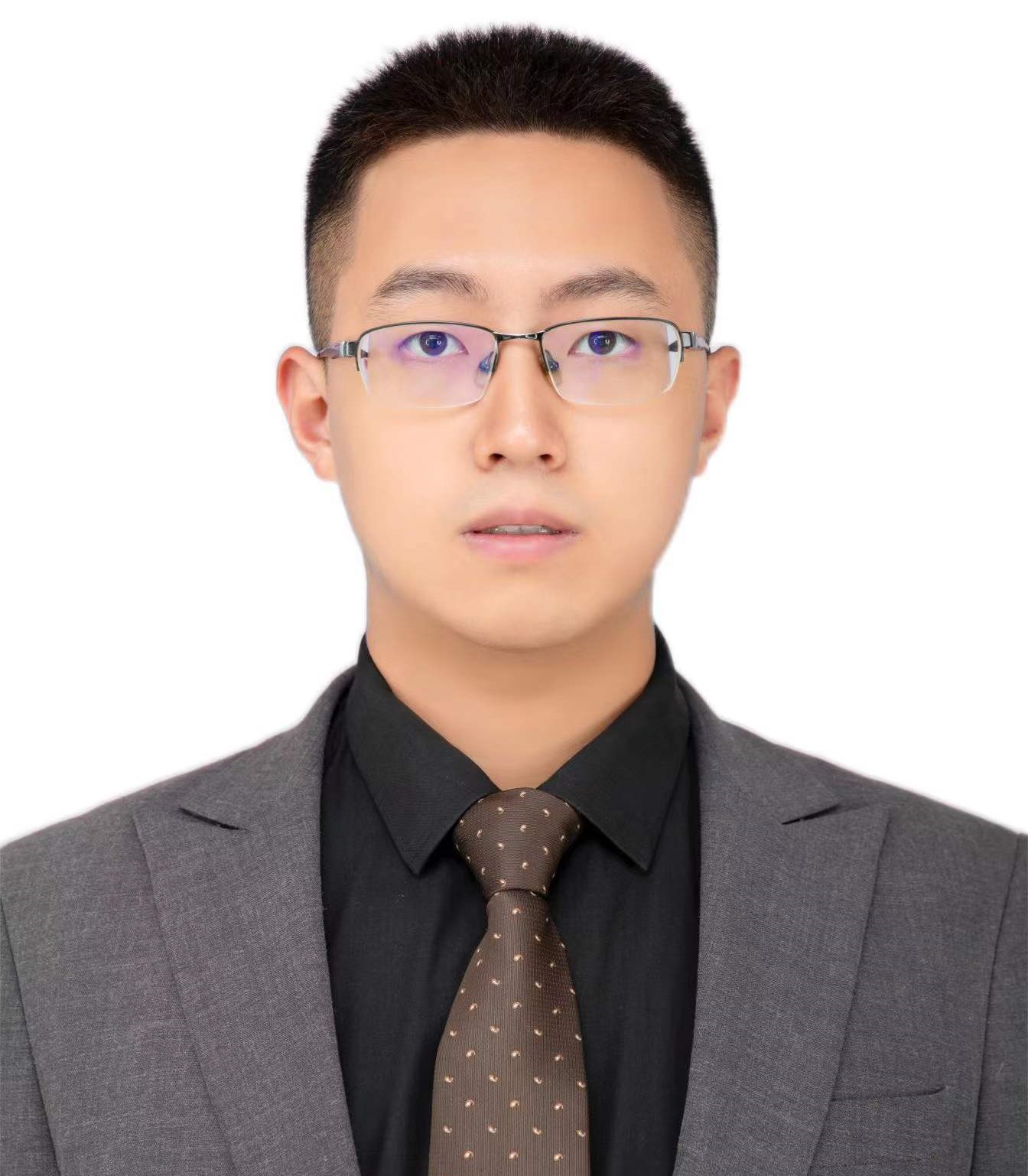}}]{Qitan Shao} received the bachelor’s degree in Intelligent Construction from the Beijing University of Technology, China, in 2021, where he is currently pursuing the master’s degree in control science and engineering. His research interest is intelligent transportation systems.
\end{IEEEbiography}

\begin{IEEEbiography}[{\includegraphics[width=1in,height=1.25in,clip,keepaspectratio]{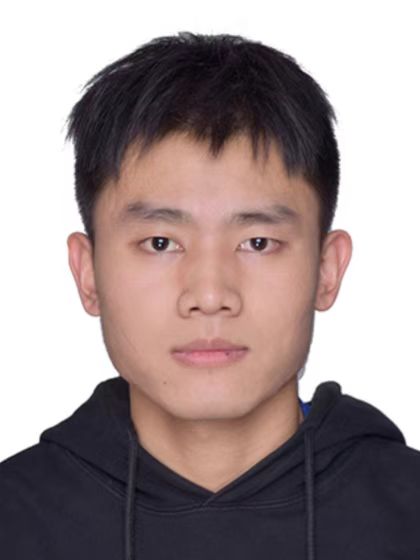}}]{Hongfei Liu} is pursuing a master degree at Beijing University of Technology. He obtained his bachelor degree in Internet of Things Engineering from Dalian Maritime University. His main research area is time series prediction and intelligent transportation systems.
\end{IEEEbiography}

\begin{IEEEbiography}[{\includegraphics[width=1in,height=1.25in,clip,keepaspectratio]{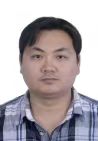}}]{Yong Zhang} (Member, IEEE) received the Ph.D.degree in computer science from the Beijing University of Technology in 2010. He is currently a Professor in computer science with the Beijing University of Technology. His research interests include intelligent transportation systems, big data analysis, visualization, and computer graphics.
\end{IEEEbiography}

\vfill

\end{document}